%% file: main.tex
\documentclass{bmvc2k}

\usepackage{wrapfig}

\title{AOL: Adaptive Online Learning for Human Trajectory Prediction in Dynamic Video Scenes}

\addauthor{Manh Huynh}{manh.huynh@ucdenver.edu}{1}
\addauthor{Gita Alaghband}{gita.alaghband@ucdenver.edu}{1}

\addinstitution{
	Department of Computer Science and Engineering \\
	University of Colorado Denver, USA
}

\runninghead{Manh Huynh, Gita Alaghband}{AOL: Adaptive Online Learning}

\begin{document}

\maketitle

\begin{abstract}
We present a novel adaptive online learning (AOL) framework to predict human movement trajectories in dynamic video scenes. Our framework learns and adapts to changes in the scene environment and generates best network weights for different scenarios. The framework can be applied to prediction models and improve their performance as it dynamically adjusts when it encounters changes in the scene and can apply the best training weights for predicting the next locations. We demonstrate this by integrating our framework with two well-known prediction models: LSTM~\cite{alahi2016social} and Future Person Location (FPL)~\cite{yagi2018future}. Furthermore, we analyze the number of network weights for optimal performance and show that we can achieve real-time with a fixed number of networks using the least recently used (LRU) strategy for maintaining the most recently trained network weights. With extensive experiments, we show that our framework increases prediction accuracies of LSTM and FPL by ~17\% and 28\% on average, and up to ~50\% for FPL on the worst case while achieving real-time (20fps).
\end{abstract}

\input{introduction}
\input{relatedwork}

\input{body}
\input{experiments}
\input{conclusion}

\bibliography{egbib}
\end{document}

%% file: introduction.tex
\section{Introduction}
\label{sec:introduction}

The problem of predicting future human movements in dynamic video scenes presents interesting challenges in developing a trajectory prediction network. Highly complex neural-based networks~\cite{alahi2016social, yagi2018future,styles2020multiple} are trained for certain scenarios first and then applied in the test phase for the purpose of predicting future locations using the best fixed network weights obtained from the extensive training. These networks learn from human past movements using feature cues such as past human locations, camera motions, human poses, and human social interactions using large datasets during the training and are expected to generalize to new testing videos/scenes. This approach often performs well for scenarios that have been similarly encountered during the training phase. However, they do not predict with high accuracy new circumstances encountered in dynamic video sequences due to sudden context changes as in camera movement, angle, speed, crowd behavior, and scenes (e.g. streets, markets, parking lots, etc.). 

To demonstrate this, one needs to look at predictions at a finer level of granularity than the average results that are normally reported. To motivate the discussion, we trained two 
\begin{wrapfigure}{l}{0.5\textwidth}
	\centering
	%\fbox{\rule{0pt}{2in} \rule{0.9\linewidth}{0pt}}
	\includegraphics[trim=0pt 12pt 0pt 0pt, width=\linewidth]{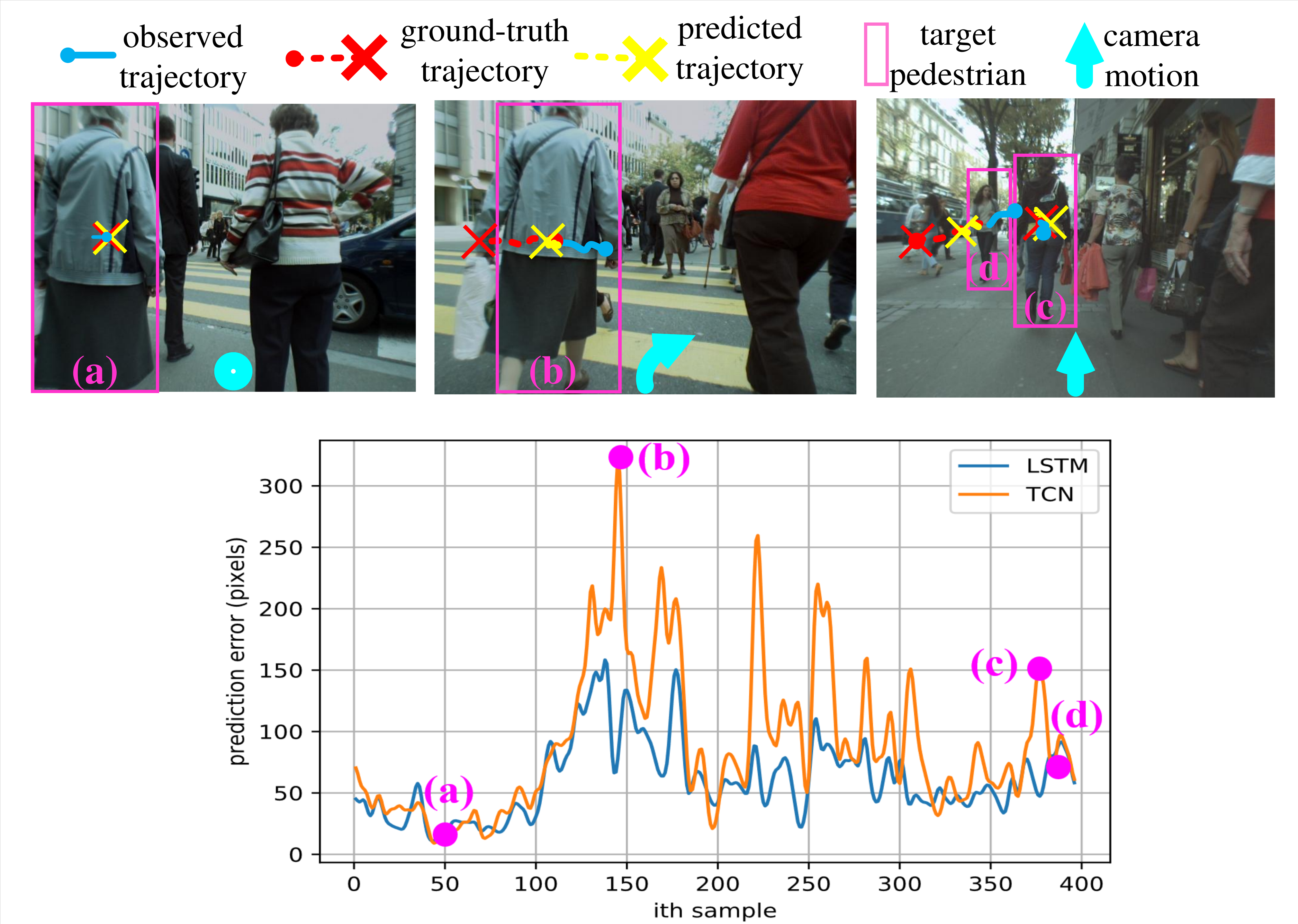}
	\caption{The trained networks (e.g. FPL~\cite{yagi2018future}, LSTM~\cite{alahi2016social}) are unable to generalize well enough when there are rapid context changes such as stable camera motions (a),  rapid sudden camera motions (b), or scenes with a mix of human movement directions (c) and (d). }
	\label{fig1}
	%\vspace{-3mm}%Put here to reduce too much white space after your table 
\end{wrapfigure}
well-known prediction networks representing different spectrum of existing methods, Long Short-term Memory (LSTM)~\cite{alahi2016social} and Future Person Location (FPL)~\cite{yagi2018future} on a large Locomotion dataset~\cite{yagi2018future} and tested them on a new video from ETH dataset~\cite{Authors06b}. 
The weights obtained at the conclusion of the training phase of each network remained fixed during testing. LSTM learns motion patterns of a person from their past positions in a sequential manner. The motion pattern encoded in LSTM hidden states can be used to predict a pedestrian’s future positions. On the other hand, FPL~\cite{yagi2018future} uses Temporal Convolutional Neural Network (TCN)~\cite{bai2018empirical} as encoders and a decoder. The encoders encode features such as human locations, scales, human poses, and camera motions extracted from observed frames. These encoded features are then concatenated (channel-wise) and decoded into predicted future locations. An advantage of using TCNs over the LSTM network is that it can encode multiple observed features in parallel, allowing more features to be incorporated with faster performance. Conversely, LSTM requires less memory and performs slightly better than TCNs for prediction in long sequences~\cite{bai2018empirical}.

Figure~\ref{fig1} shows the future location prediction results in final displacement error in pixels (FDE), where each sample corresponds to a pedestrian’s future trajectory in the next one second or 10 frames.  From the variations plotted, we observe that neither network can perform consistently well across all samples. While both networks predict with high accuracy (lower prediction error) when the camera motions are stable (scenario a); they perform poorly when there are abrupt camera motions (scenario b). Another context change that the two networks do not capture well is shown in scenarios c and d when the camera is steady (camera consistently moves forward), predicting the future locations of pedestrians moving-forward (scenario c) is harder than pedestrians moving-away (scenario d).  

To address this problem, we believe that a network should have the ability to adapt its weights to different contexts. The first approach we explore in this paper, base adaptive online learning framework (B-AOL), is to continue having the ability to train after the initial training phase. B-AOL consists of a master network and a slave network. Both share the same prediction network architecture such as LSTM~\cite{alahi2016social} or FPL~\cite{yagi2018future}. While the master network continuously trains and its weights change as it encounters new contexts, the slave network uses a copy of master’s pre-trained weights (last training stage) to test the incoming samples. The weights for the slave network are switched to that of the master network upon encountering new contexts where the slave predictions are inaccurate. This process ensures that the master network is always updated with contexts of recent samples to achieve higher prediction accuracy for the next samples as there is a high probability that nearby samples possess similar temporal dynamics. However, this approach can fail when several abrupt context changes are encountered close to each other. In such a case, B-AOL will not be able to capture all scenarios due to inadequate training data. 
We address this issue by designing an adaptive online learning model (AOL) to efficiently adapt to dynamic scene context changes by generating and maintaining one master network that continues to train, and multiple slave networks, each with the best trained set of weights matching a specific context encountered during the testing process up to the current point.  This allows us to cope with the context changes in the next testing samples as there is a high probability that the new sample contexts will be similar to one of our previously generated best-performing network weights. The challenge arises because the number of prediction network weights increases linearly with the number of new contexts encountered in the streaming video resulting in large memory usage and slow execution time. To handle this problem, we maintain a Least Recently Used (LRU) network replacement policy where we can control the number of slave networks. The LRU enables us to keep the most updated slave networks in the list by continuously updating their weights with recent master network’s weights replacing those that have not performed well (used) for the longest time. More importantly, by controlling the number of networks, our framework is able to perform in real-time.

Our adaptive online learning framework is flexible in that it can incorporate different types of prediction networks and improve their prediction accuracy (Section~\ref{sec:aol}). We conduct extensive experiments integrating our framework with two state-of-the-art prediction network models on two dynamic video sequence datasets: First-person Locomotion~\cite{yagi2018future} and ETH~\cite{ess2008mobile} and show significant accuracy improvements over the stand-alone prediction networks on these datasets (Section~\ref{sec:experiements}).

%% file: relatedwork.tex
\section{Related Work}
\label{sec:relatedwork}
\textbf{Human Trajectory Prediction in Dynamic Video Scenes.} Most of the recent work~\cite{alahi2016social,yagi2018future,voigtlaender2017online} rely on recurrent neural networks (RNNs)~\cite{mikolov2010recurrent}, temporal convolutional neural networks (TCNs)~\cite{bai2018empirical}, or their variants~\cite{gupta2018social,zhang2019sr} to predict human future locations. Furthermore, several attempts have been made to model human-scene interactions~\cite{bartoli2018context,huynh2019trajectory}and/or human-human interactions~\cite{gupta2018social,yagi2018future}. However, the aforementioned methods works only perform well in static scenes, where bird’s eye views are given. Forecasting human future location in dynamic scenes is more challenging due to the abrupt camera motions, vast varieties of human poses and dynamic scene change from one place to another. Some~\cite{yagi2018future,styles2020multiple} have attempted to model the camera motions and human poses in prediction networks. They focus on improving network architectures and crafting input features. However, with the limited training data, these networks do not generalize in different contexts during testing. To tackle this problem, we propose a novel adaptive online learning framework to boost the accuracy of the existing prediction models. \\
\textbf{Online Network Adaptation.} To the best of our knowledge, there is no existing online adaptation work for trajectory prediction in dynamic scenes. The recent research in unsupervised domain adaptation~\cite{ganin2014unsupervised} and meta-learning~\cite{finn2017model} possibly presents the closest ideas to our work, but they are fundamentally different. Meta-learning~\cite{finn2017model} presents the a concept of learning-to-learn, which learns specific model parameters for a given domain. However, a small ground-truth data for each domain must be given, which limits its applicability to real-time and practical applications. Common approaches~\cite{ganin2014unsupervised, tasar2020colormapgan, gholami2020unsupervised} for unsupervised domain adaptation minimize domain discrepancy between source and target domain from raw input data without the need  for ground truth data in target domain. Though, metadata (e.g. weather conditions, day/night,…) must be available. Due to this assumptions, it is non-trivial to apply the methods in meta-learning and unsupervised domain adaptation for human trajectory prediction in dynamic scenes. There are some online adaptation methods, which are customized for applications such as object tracking~\cite{park2018meta}, video segmentation~\cite{voigtlaender2017online}, robot motion planning~\cite{nagabandi2018deep}. But, it is not straightforward to adapt these networks for predicting human future locations due to network architecture differences.

%% file: body.tex
%\section{Adaptive Online Learning Framework (AOL) for Human Future Movement Prediction}
%\label{sec:body}
%In this section, we first present our base online adaptation framework (B-AOL) and discuss its shortcomings. We then describe the design of AOL. The LSTM and FLP prediction networks and their input features that are used in our frameworks are detailed last.
\section{B-AOL: Base Online Adaptation Framework.}
\label{sec:baol}
\begin{figure}[!]
	\centering
	%\fbox{\rule{0pt}{2in} \rule{.9\linewidth}{0pt}}
	\includegraphics[width=0.9\linewidth]{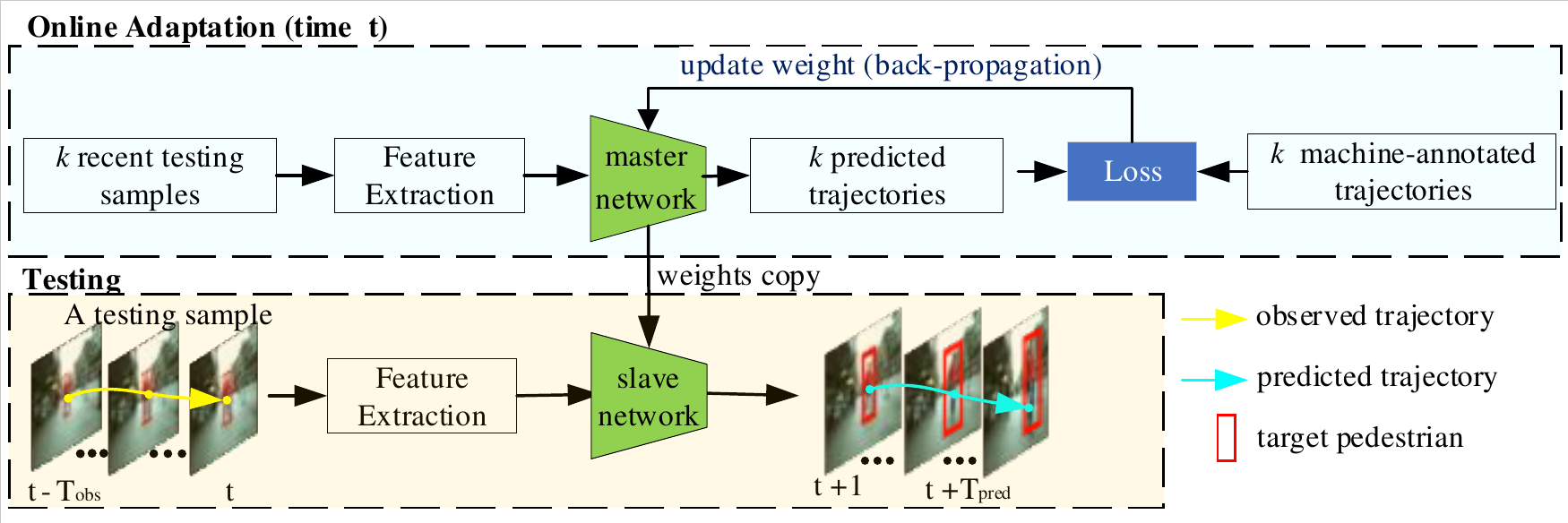}
	
	\caption{Base online adaptation framework (B-AOL) for human trajectory prediction. }
	\label{fig2}
	%\vspace{-5mm}%Put here to reduce too much white space after your table 
\end{figure}
\begin{wrapfigure}{l}{0.5\textwidth}
	\centering
	%\fbox{\rule{0pt}{2in} \rule{0.9\linewidth}{0pt}}
	\includegraphics[trim=0pt 0pt 0pt 0pt, width=0.9\linewidth]{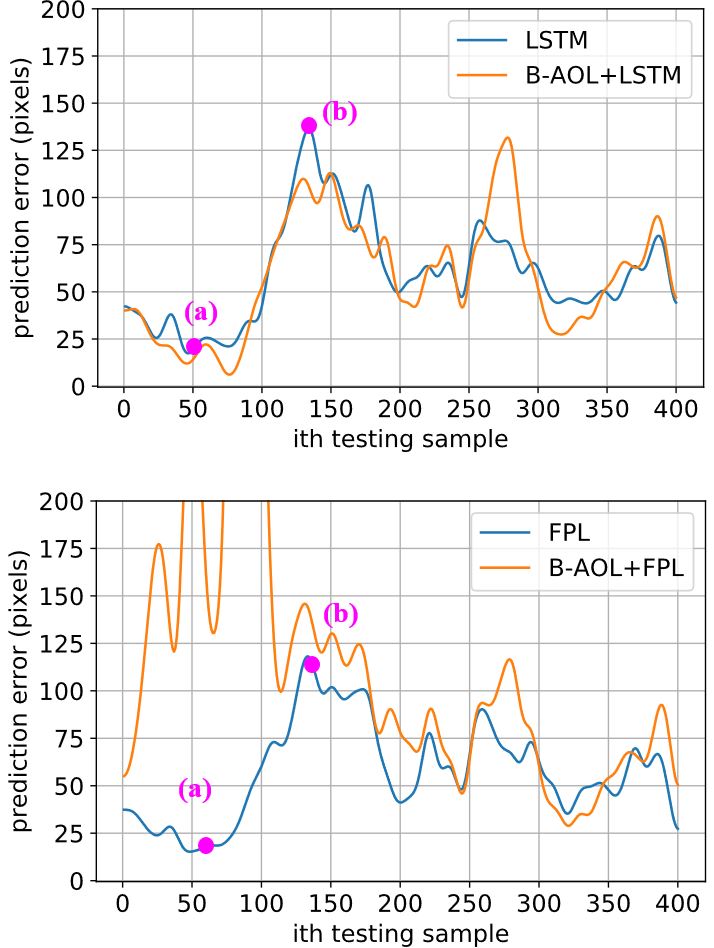}
	\caption{The prediction error (lower is better) of base online adaption framework (B-AOL) (with k =1) compared with no-adaptation networks on each testing sample using LSTM network (top graph) using FPL network (bottom graph).}
	\label{fig3}
	%\vspace{-3mm}%Put here to reduce too much white space after your table 
\end{wrapfigure}
The base online adaption framework (B-AOL) adapts a prediction network (LSTM and FLP in this paper) to the context changes after an initial training phase by updating the network’s weights using k recent testing samples. Figure~\ref{fig2} depicts the overview of B-AOL for human trajectory prediction in dynamic streaming video. The framework consists of a master network and a slave network. 

In the online adaptation phase, the master network’s weights are first initialized with the best-trained prediction network’ s weights on a large dataset. At each time $t$ of a new video being tested, the master network is trained using $k$ recent testing samples. To train the master network, we calculate a mean square error (MSE) loss between the recently predicted trajectories and the high confident machine-annotated trajectories generated from a human tracking algorithm (e.g. Track-RCNN~\cite{voigtlaender2017online}. The loss’s gradient is backpropagated to update the master network’s weights using mini-batch gradient descent~\cite{ruder2016overview}. Its trained weights are then copied to the slave network for testing the upcoming samples. Note that using the high confident machine-annotated data does not cause any error drifts during training and has been efficiently used as augmented data \cite{voigtlaender2017online}.

 In the testing phase (prediction phase), the slave network predicts future locations of the target pedestrian (in the red bounding box) in the next $T_{pred}$ frames by observing the past human trajectories and other features extracted from $T_{obs}$ frames. For evaluation, the predicted future locations are compared (using MSE) with ground-truth trajectories. The details of feature extraction and evaluation metrics are discussed in Section~\ref{sec:experiements}. The base online adaptation framework poses two sets of challenges: 

(1) B-AOL cannot effectively adapt to sudden and unexpected changes in different scene contexts. Figure~\ref{fig3} shows examples of this. We integrated the LSTM and FLP prediction networks with our B-AOL framework and repeated the experiment presented in Section~\ref{sec:introduction} (B-AOL+LSTM and B-AOL+FPL). The graphs show FPL performs better than LSTM in general; but also that B-AOL+LSTM and B-AOL+FPL do not show improvements in either case when there are abrupt camera motions (Figure~\ref{fig3} , scenario b). Additionally, we observe that B-AOL+FPL’s prediction accuracy worsens compared to FPL even when the camera motion is stable at the beginning of testing video due to the domain shift problem~\cite{sankaranarayanan2018learning}. This usually happens when the data (features distributions from the training datasets are different from the test datasets.

(2) To improve the prediction accuracy of B-AOL, one can increase the number of k recent testing samples and use a higher number of training epochs. However, due to the runtime constraints, this approach is unrealistic for real-time applications. The study of how sample sizes and training epochs impact the prediction accuracy and processing time are presented in the ablation studies in Section~\ref{sec:experiements}. 
\vspace{-3mm}%Put here to reduce too much white space after your table 
\section{AOL: Adaptive Online Learning Model.}
\label{sec:aol}
To address the shortcomings of the B-AOL framework, we propose a novel adaptive online learning framework (AOL). The key idea is to maintain a master and a list  of multiple slave prediction networks $S=\{s_0,s_1…,s_n\}$, where $s_i$ performs best in a specific context $i$. Having $n+1$ slave networks enables AOL to predict with higher accuracy for $n+1$ different scene contexts. Thus, when the new scene contexts are encountered, there is a probability that the new scene context will be similar to one of the previously learned $n+1$ contexts.

\begin{figure}[t]
	\centering
	%\fbox{\rule{0pt}{2in} \rule{.9\linewidth}{0pt}}
	\includegraphics[width=0.9\linewidth]{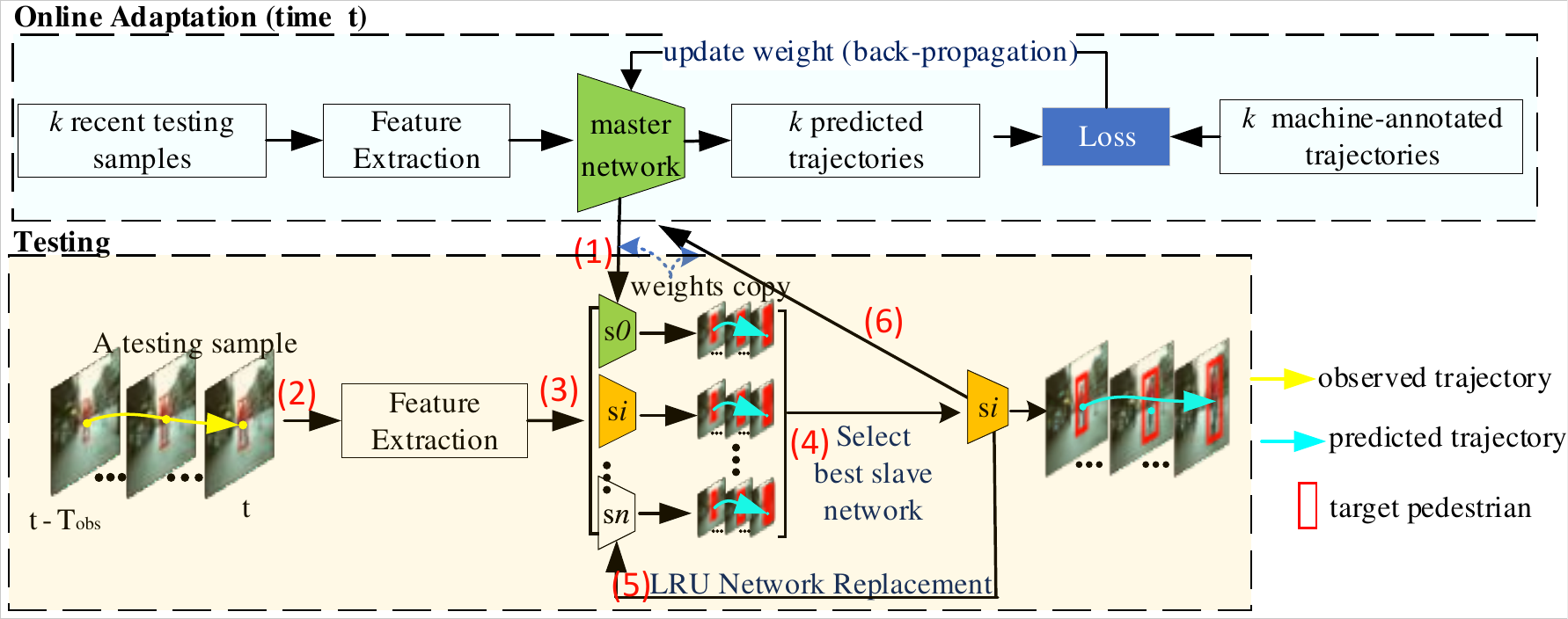}
	
	\caption{The overview of the proposed framework (AOL) for human future trajectory prediction.}
	\label{fig4}
%	\vspace{-5mm}%Put here to reduce too much white space after your table 
\end{figure}

The AOL framework is depicted in Figure~\ref{fig4}. We use the same online adaptation procedure as B-AOL for training the master network. The testing process works as follows:

\begin{itemize}
	\item 	At time  $t= 0$, we initialize the slave network list $S$ with $s_0$, $S \leftarrow \{s_0\}$. At every time step, the trained master network’s weights obtained from the recent testing samples are copied to the slave network $s_0$ (Figure~\ref{fig4}, step 1). This ensures we always have a slave network (i.e. $s_0$) that is updated with the most recent context changes. 
	\item To predict a target’s future locations in the next $T_{pred}$ frames, we extract features from $T_{obs}$ frames. The extracted features are then input to all slave networks ${s_0,s_1…,s_n}$ (Figure~\ref{fig4}, Step 2) to generate a list of predicted trajectories corresponding to the output of the slave networks. We note that the slave networks share the same network architectures, only their weights are different.  The slave network that generates the best predicted trajectory is selected. 
	\item To find the best predicted trajectory (Figure~\ref{fig4}, step 4), we calculate the mean square error (MSE) between all predicted trajectories in the list with the machine-annotated trajectories generated from a human tracking algorithm (e.g. Track-RCNN~\cite{voigtlaender2017online}). The best predicted trajectory is the one with the lowest MSE. Using machine-annotated trajectories is appropriate in testing phase because one should not have access to ground-truth trajectories while testing. 	
	\item Lastly, the list of slave networks is updated using the following scenarios: (1) If the best-selected slave network $s_b= s_i$ with $i>0$ then use prediction from this slave network. (2) If the best-selected slave network is $s_0$ (i.e. $s_b=s_0$). We save a copy of $s_b$ into the slave network list using Least Recently Used (LRU) network replacement policy. Specifically, if the current number of slave networks $p$ has not reached the pre-defined maximum capacity ($p<n$), we add this network as a new entry in the list: $s_{p+1} \leftarrow copy(s_0) \label{eq1}; 	S \leftarrow  S \cup s_{p+1} \label{eq2}$. 
	Otherwise, we replace it with the least recently used (LRU) network in the slave network list. Using LRU not only we maintain slave networks with the most recent contexts, we also can control the total number of these networks and achieve real-time processing.\\
\end{itemize}

\vspace{-5mm}%Put here to reduce too much white space after your table 

%% file: experiments.tex
\vspace{-10mm}%Put here to reduce too much white space after your table 
\section{Experiments}
\label{sec:experiements}
\subsection{Experiment Setups}
\textbf{Datasets.} We use Locomotion~\cite{yagi2018future} and ETH~\cite{ess2008mobile} datasets for evaluation. The Locomotion dataset is captured from pedestrians wearing a chest-mounted camera on busy streets in Japan. The dataset consists of ~$50,000$ samples of ~$5,000$ pedestrians in total. Each sample consists of  $T= 20$ frames ($10$ frames for observation and 10 frames for prediction). \\
We also conduct experiments on ETH datasets~\cite{ess2008mobile}, captured from robots’ cameras. The ETH datasets consist of 4 video sequences with $~9000$ samples are extracted. 
All frames in both datasets are scaled to size $960x1280$ and frame per second (fps) is $20$.  \\
\textbf{Evaluation Metrics.} We evaluate our system using two commonly used metrics~\cite{yagi2018future,alahi2016social,styles2020multiple}: (a) average displacement error (ADE): mean square error over all locations of predicted and true trajectories; (b) final displacement error (FDE): mean square error at the final predicted and true locations of all human trajectories.\\
\textbf{Feature extraction.} We used the following process to extract the required features for FPL and LSTM. For each video sequence frame, we extract an $18$ key-point pose (36-dimensional vector) per pedestrian using OpenPose~\cite{cao2018openpose}. The human location (2-dimensional vector) is set at the middle hip of the pose, the body scale (1-dimensional vector) is calculated as the height of the human body from the neck to the middle hip. To calculate the camera motions (i.e. ego-motions), we use grid optical flows. First, the optical flow for each pixel is calculated using FlowNet2~\cite{ilg2017flownet}. We then divide a frame into $3x4$ grid-cells and calculate the average optical flow of all pixels within a grid-cell. The resulting 24-dimensional optical flow vector is used to represent the camera motion at a given frame. During the pre-training process of a prediction network dataset, we use ground-truth human trajectories for calculating the prediction loss. While during testing, we use the machine-annotated trajectories.
\begin{wraptable}{r}{0.5\textwidth}
	
	\includegraphics[width=\linewidth]{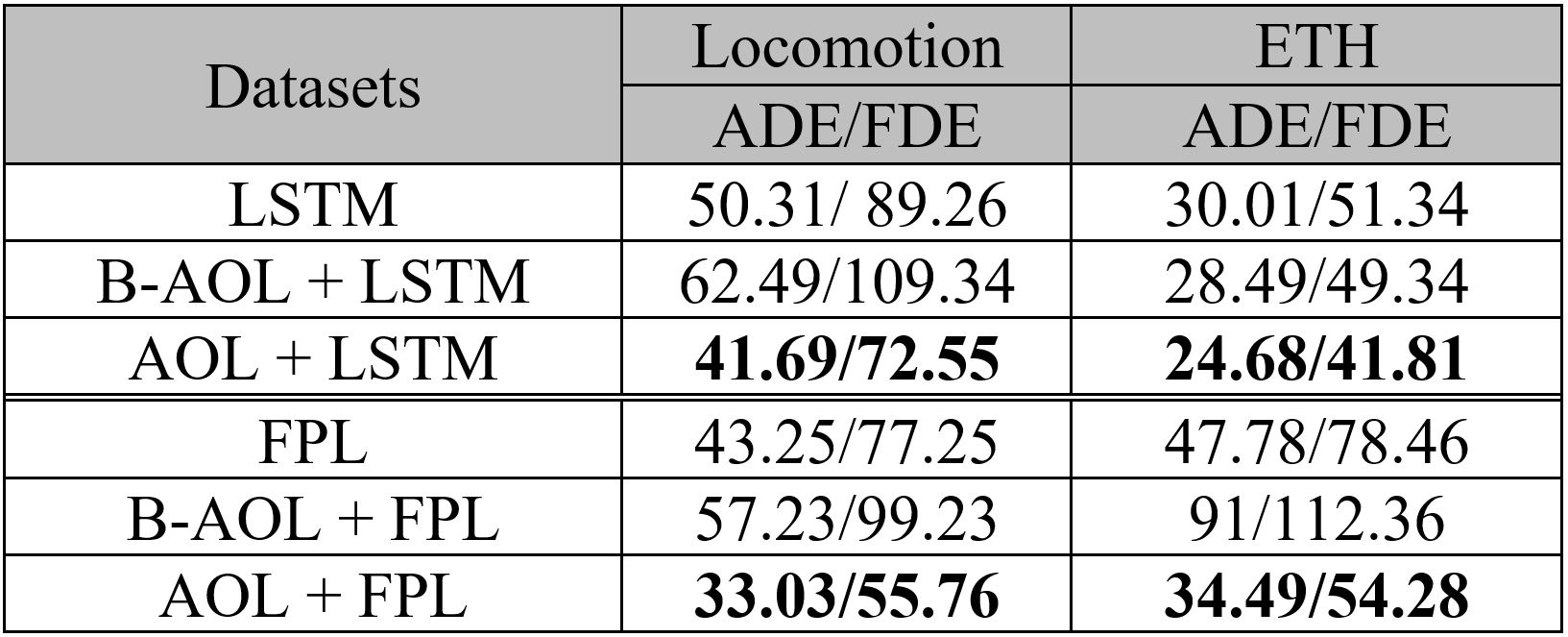}
	\caption{Quantitative results of AOL on two datasets: Locomotion and ETH.}
	\label{table1}
	
	\medskip\par
	\includegraphics[width=\linewidth]{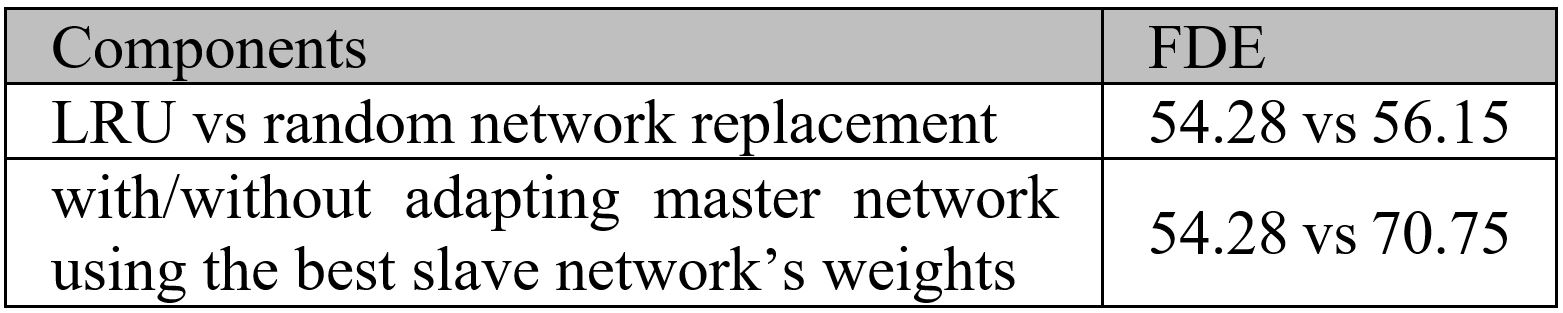}
	\caption{Contributions of framework components.}
	\label{table2}
	%\vspace{-3mm}%Put here to reduce too much white space after your table 
\end{wraptable}
\textbf{Implementation details}. We implemented our framework using PyTorch~\cite{ketkar2017introduction} deep learning framework. During the online adaptation phase, the number of training epochs is set to 3, the number of recent samples $k=1$, the learning rate is $0.001$. The maximum number of slave networks is 10. The analysis of these parameters is presented in Section~\ref{sec:analysis}. We use Adam~\cite{ilg2017flownet} for network optimization. To generate the machine-annotated trajectory of each sample, we used Track-RCNN~\cite{voigtlaender2017online}. The network models are trained/tested on GPU Tesla P100-SXM2.\\
\vspace{-5mm}%Put here to reduce too much white space after your table 
\subsection{Quantitative results on Locomotion and ETH datasets}
\textbf{Results on the Locomotion dataset.} We apply the 5-fold cross-validation protocol similar to~\cite{yagi2018future}. The dataset is split into 5; the prediction networks (LSTM and FPL) are initially trained on four splits (pre-training phase); we then test/adapt them using AOL on the remaining split. The average ADE/FDE over 5 splits are reported in Table~\ref{table1}. Our framework AOL reduces the ADE/FDE of LSTM and FPL by $17.13\%/12.34\%$ and $28.25\%/27.81\%$, respectively. We can see that FPL is a better prediction network for adaptation. With more input features used, stand-alone FPL also performs better than LSTM (43.25 vs 50.31 ADE). However, B-AOL worsens the prediction results of both LSTM and FPL models as expected. \\
\textbf{Results on the ETH dataset.} We tested AOL on a different dataset to demonstrates it can predict with higher accuracy in dynamic scenes and effectively adapt to domain changes. We train LSTM and FPL on the Locomotion dataset using 100 training epochs and continue adapting/testing it using AOL on the ETH dataset. The second column in Table~\ref{table1} shows AOL improves ADE/FDE of LSTM and FPL by $17.76\%/18.56\%$ and $27.78\%/30.81\%$ respectively. Interestingly, we notice that FPL does not perform better than LSTM on ETH datasets. This is due to the missing human pose key-points of several pedestrians in the ETH videos. This scenario happens frequently for pedestrians up-close to the camera or too small when too far from the camera. However, AOL still effectively adapts FPL by improving its prediction accuracy in these scenarios.  
\subsection{Analysis}
\label{sec:analysis}
In this section, we study in-depth the impacts of our design ideas and other parameters that impact on AOL’s performances. The experiments are done using the ETH dataset.\\
\begin{figure}
	\centering
	%\fbox{\rule{0pt}{2in} \rule{.9\linewidth}{0pt}}
	\includegraphics[width=\linewidth]{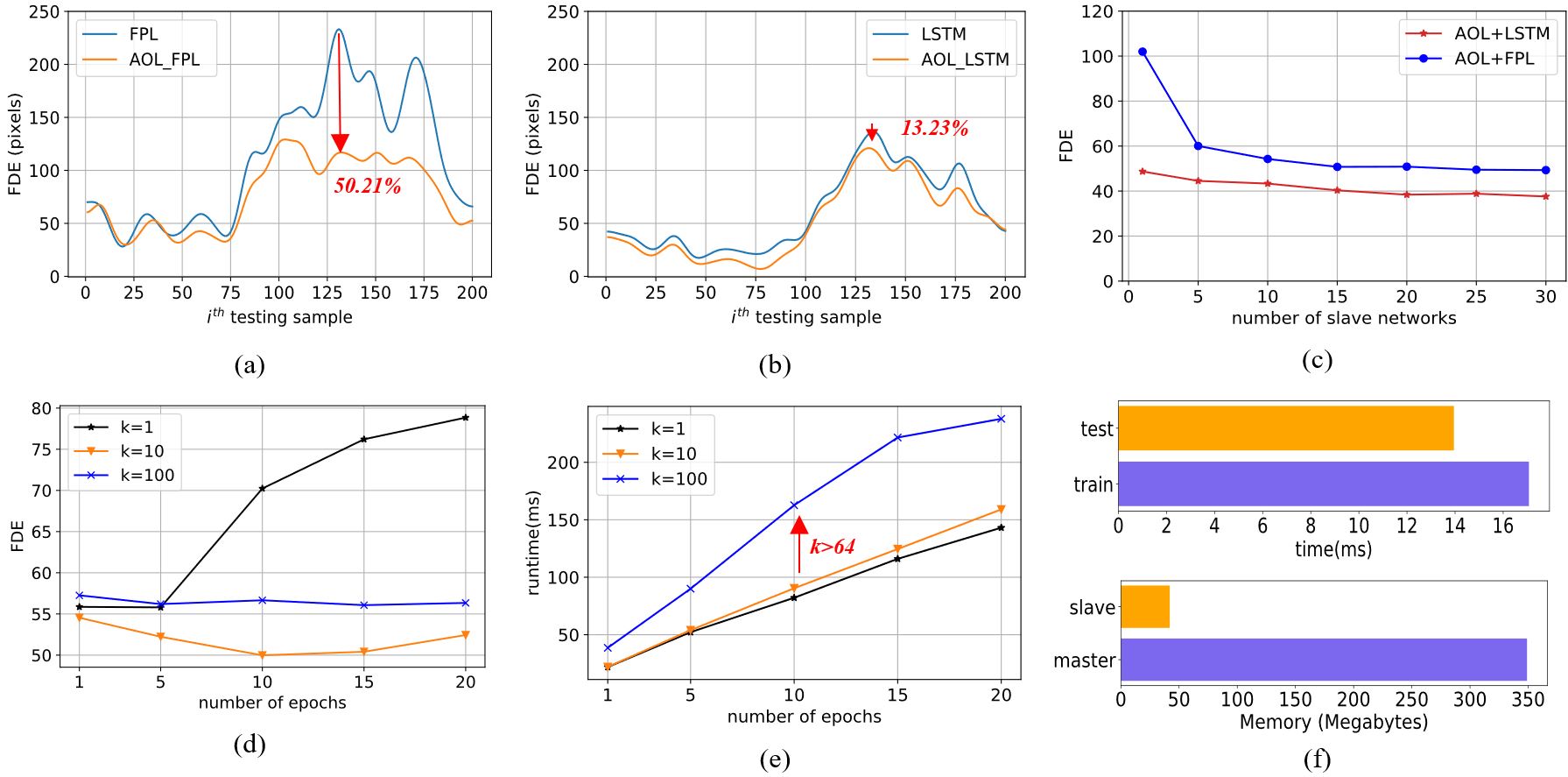}
	
	\caption{Analysis. (a,b) efficiencies of AOL under abrupt camera motion; (c) impact of number of slave networks; (d, e) impact of number of past samples and training epochs on FDE and runtime; (e) runtime and space complexities.}
	\label{fig5}
	%\vspace{-5mm}%Put here to reduce too much white space after your table 
\end{figure}
\textbf{Contributions of framework components.} Table~\ref{table2} shows the impacts of two design ideas of AOL: (a) using the least recently used (LRU) network replacement policy, and (b) copying the best slave network’s weights to the master network for adapting the prediction/training to encounter new scene contexts. Compared to random network replacement policy, LRU maintains the most recent contexts. As a result, it generates more accurate predictions (lower FDE) compared to random network replacement (54.28 vs 56.15). Copying the best slave network’s weights to the master network, so that the training can continue for the best and most recent prediction, also produces significantly higher prediction accuracy (54.28 vs 70.75 FDE). This is due to the similar temporal dynamics of samples in nearby frames. \\
\textbf{Prediction under abrupt camera motions.} Figures~\ref{fig5}a and~\ref{fig5}b show that AOL successfully handles abrupt camera motion scenarios. We observe that AOL reduces the prediction error (FDE) by up to 50.21\% for FPL and 13.23\% for LSTM for the worst-case scenario. \\
\textbf{Impact of varying number of slave networks.}  Figure~\ref{fig5}c shows the resulting FDE as we change the number of slave networks. As we increase the number of networks, FDE is consistently reduced until about 10 networks and seems to saturate when the number of networks reaches 20. \\
\textbf{Time complexity.} Figure~\ref{fig5}f (top bar) shows the runtime of testing and training of the AOL+FPL given one testing sample and 10 slave networks. With only one sample used for training and 3 training epochs, the training time is ~17 milliseconds ($ms$) per sample; while the test time of 10 slave networks is $13.91ms$ per sample.  If the slave networks are tested in parallel, the testing time can be mitigated (to about $1.3ms$). Overall, the total processing time per frame is ~$31ms$ sequentially, which is real-time processing in a video of framerate 20fps.\\
\textbf{Space complexity.} a master FPL network requires about 350 Megabytes (MBs), while each slave network requires a factor 7 less memory (Figure~\ref{fig5}f, bottom bar). This is because the master network must store additional network parameters (i.e. gradients), while the slave network does not. However, the more slave networks used, the more memory is needed. \\
\textbf{Impact of varying past samples ($k$) and training epochs($e$)}. Figures~\ref{fig5}d and ~\ref{fig5}e show FDE and runtime as we change the number of recent samples and training epochs. We fixed the number of networks to 10 and use the FPL prediction network. We note that when $k$ is small, $e$ (epoch size) should also be small. Otherwise, it is very easy to overfit to a specific context. On the other hand, when $k$ is large, FDE does not improve because each slave network has an average performance in many contexts. Larger $k$ and $e$ also significantly affect the runtime (Figure~\ref{fig5}e). Increasing $k$ and $e$ linearly increase the runtime. Interestingly, with batch-train size 64, we see a big jump in the runtime for $k=100$. Thus, using large samples and training epochs are impractical for this real-time application. With 10 slave networks, we need to keep e smaller than 5 and $k$ smaller than batch size to achieve real-time processing.
\subsection{Qualitative results.}
\begin{figure}
	\centering 
	
	%\fbox{\rule{0pt}{2in} \rule{.9\linewidth}{0pt}}
	\includegraphics[width=1\linewidth]{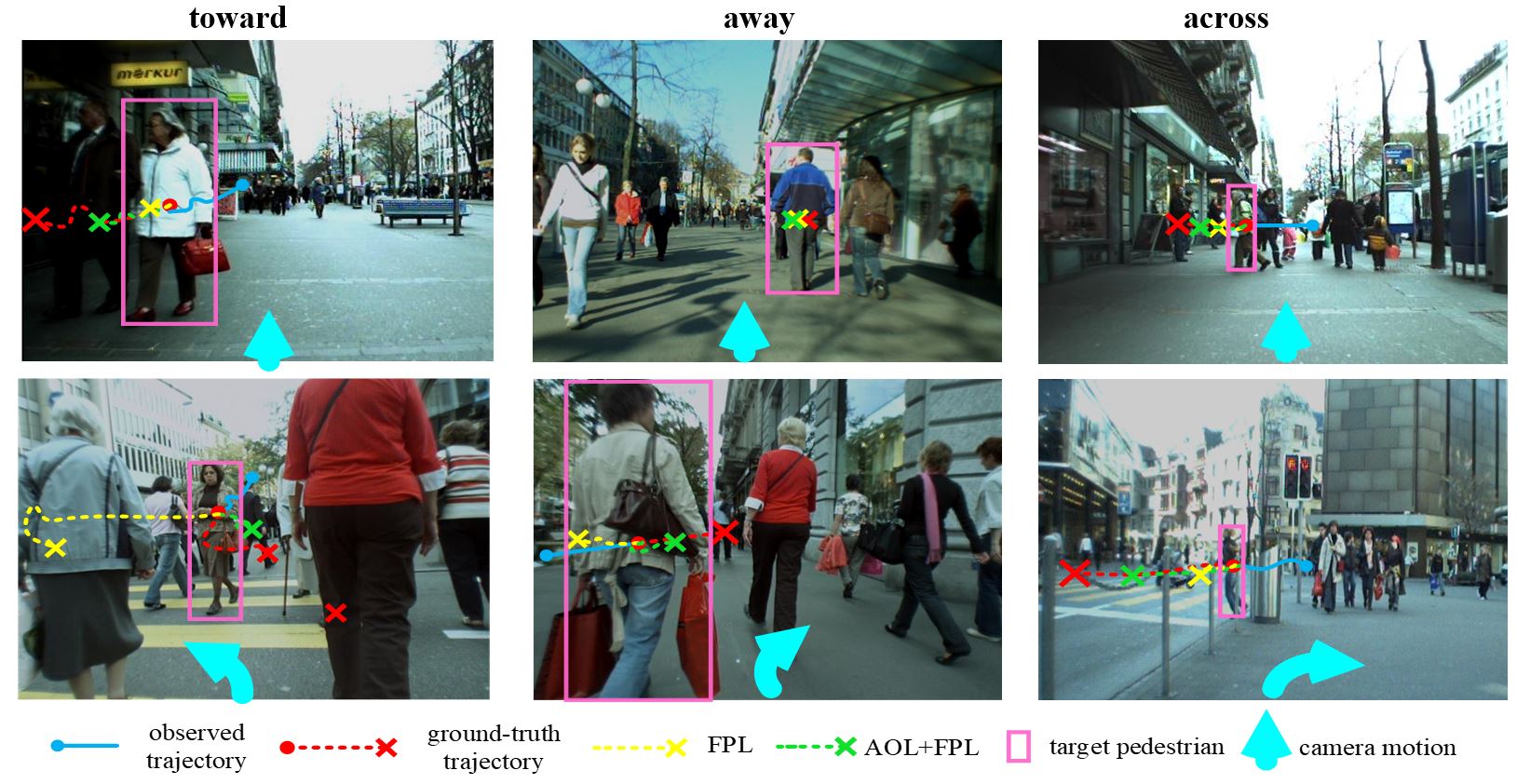}
	
	\caption{Qualitative results. AOL significantly improves the performance of FPL under various pedestrian movements and camera motions. Camera motions are stable on top figures, while there are abrupt changes in the bottom figures}
	\label{fig6}
	%\vspace{-5mm}%Put here to reduce too much white space after your table 
\end{figure}

Figure~\ref{fig6} presents sample qualitative results showing improvements of AOL+FPL over stand-alone FPL on ETH datasets under various pedestrian movements: moving toward, away, across under different camera motions: stable and abrupt. When the camera is stable (top row), FPL and AOL+FPL produce comparable results. However, when there are abrupt camera motions, FPL falls short, while AOL+FPL maintains better prediction results.

%% file: conclusion.tex
\section{Conclusions}
\label{sec:conclusions}
In this paper, we presented a novel adaptive online learning framework (AOL) for human future location prediction in dynamic video scenes. AOL relies on the idea of a master network, which continuously trains to keep up with changes in scene contexts, and multiple slave networks, capable to produce the highly accurate predictions for the most recent n scene contexts encountered. AOL uses LRU to control the number of slave networks to achieve real-time and limit memory consumptions. AOL can integrate prediction network models and improve their performance in dynamic scenes. We presented this by integrating two well-known LSTM and FLP prediction networks with AOL and showed high adaptability and significant improvements in the prediction accuracy of these networks while achieving real-time performance.